\documentclass[sigconf]{acmart}
\AtBeginDocument{%
  \providecommand\BibTeX{{%
    \normalfont B\kern-0.5em{\scshape i\kern-0.25em b}\kern-0.8em\TeX}}}

\copyrightyear{2022}
\acmYear{2022}
\setcopyright{rightsretained}
\acmConference[CBMI 2022]{International Conference on Content-based Multimedia Indexing}{September 14--16, 2022}{Graz, Austria}
\acmBooktitle{International Conference on Content-based Multimedia Indexing (CBMI 2022), September 14--16, 2022, Graz, Austria}
\acmDOI{10.1145/3549555.3549592}
\acmISBN{978-1-4503-9720-9/22/09}

\usepackage[shortlabels]{enumitem}
\AtBeginDocument{%
  \providecommand\BibTeX{{%
    \normalfont B\kern-0.5em{\scshape i\kern-0.25em b}\kern-0.8em\TeX}}}
\usepackage{hyperref}
\usepackage{array,xcolor,colortbl}
\usepackage{algorithm}
\usepackage{url}
\usepackage{multirow}
\usepackage{amsmath} 
\usepackage{graphics}
\usepackage{subfigure}
\newcolumntype{G}{>{\centering\arraybackslash\columncolor{gray!20!white}}p{0.075\textwidth}}

\begin{document}

\title{Analysing the  Memorability of a Procedural Crime-Drama TV Series, CSI}

\author{Seán Cummins, Lorin Sweeney and Alan F. Smeaton}
\email{Alan.Smeaton@DCU.ie}

\affiliation{%
  \institution{Insight SFI Research Centre for Data Analytics,\\Dublin City University, Glasnevin, Dublin 9}
  \country{Ireland}
}


\begin{abstract}
 We investigate the  memorability of a 5-season span of a popular crime-drama TV series, CSI,  through the application of a vision transformer fine-tuned on the task of predicting video memorability. By investigating the popular genre of crime-drama TV through the use of a detailed annotated corpus combined with video memorability scores, we show how to extrapolate meaning from the memorability scores generated on  video shots. We perform a quantitative analysis to relate video shot memorability to a variety of aspects of the show. The insights we present in this paper illustrate  the importance of video memorability in  applications which use multimedia in areas like 
 education, marketing, indexing, as well as in the case here namely TV and film production. 
\end{abstract}




\maketitle

\section{Introduction}
\label{sec:introduction}
Memorability is a subconscious measure of importance imposed on external stimuli by our internal cognition which ultimately decides the experiences we either carry with us into our long-term memory, or we discard. Memory is an unequivocally unreliable phenomenon for any person and despite its importance, it is challenging for us to influence  what we will ultimately remember or forget. This absence of any unified meta-cognitive insight brings meaning to the study of memory, or more specifically into the study of memorability, more generally known as the likelihood that something will be remembered or forgotten \cite{Sweeney2021}.

We have evolved to store the important moments in life while forgetting uninteresting or redundant information. Thus, a memorability-based model of our experiences could allow us to filter and sort information we encounter using a human-memorability criterion \cite{Newman2020}. Researching how our memories take shape could have a profound impact on a plethora of research areas such as human health or the curation of educational content \cite{Mediaeval2020,Mediaeval2021}. 

The study of memorability may also have a commercial impact in fields such as marketing, film or TV production, or content retrieval \cite{Yue2021}. The latter of these use-cases serves as a particularly important application of memorability as the volume of digital media content available grows exponentially over time via platforms such as social networks, search engines, and recommender systems \cite{Shekhar2017,Siarohin2019,Newman2020}.

In this research, crime-drama exemplified in television (TV) programs such as \textit{CSI: Crime Scene Investigation} (referred to as CSI) acts as an experimental test bed in which computational memorability as a surrogate for cognitive processing, can be analysed. CSI is a procedural crime-drama TV series with each episode following the same format. Viewers are introduced to a crime at the beginning of an episode. As the episode progresses, various clues are revealed, concluding by revealing the true perpetrator of the crime. The success of CSI can be attributed partly to its  aesthetic high production values which many media critics comment on, and we argue its success is also due to its ability to promote critical thinking and memory in its viewers. 

Previous studies have already proven crime-drama TV series to be useful as test cases in multiple research areas including NLP, summarisation and multi-modal inference \cite{Frermann_Cohen_Lapata_2018,Papasarantopoulos2019,Papalampidi2020}.
In particular, the work in \cite{Salt2006} is a large-scale anthology of Barry Salt's essays on statistical analysis of film over many years of investigation. Salt has single-handedly established statistical style analysis as a research paradigm in film studies. He argues that ``many directors have sharply different styles that are easily recognized'' \cite[p.~13]{Salt2006}. 
Salt's more recent work  includes  the application of film theory to the creation of TV series and in \cite{doi:10.1386/jmpr.2.2.98} he presents a stylistic analysis of episodes in twenty TV drama series, finding that the structure of  TV style is uniform, a similar finding to movies.

The work by Salt and others uses a variety of measures such as shot duration and shot distribution 
and since then several further studies have attempted to perform statistical analysis of TV content \cite{Schaefer2009,Butler2014,Redfern2014,Arnold2019,Kim2020}.
Yet none of this, or any other, forms of analysis considers the memorability of the visual content in its analysis.  

We investigate the distribution of computational memorability scores of CSI episodes over 5 seasons by leveraging modern vision transformer  architectures \cite{Dosovitskiy2021,Bao2021,Radford2021} fine-tuned on the task of media memorability prediction. We interleave our predicted memorability scores for given episodes of CSI with a detailed annotated corpus to investigate correlation between memorability and aspects of the show including scene and character significance.

Our research investigates whether there is a relationship between the memorability of a shot/scene and the characters present in it, how the memorability associated with particular characters develops over multiple episodes and seasons, and whether the memorability of a shot/scene is correlated with the significance of the scene.
The significance of this and the reason we study the CSI TV series is precisely because of its importance in popular culture. Since its first broadcast in the early 2000s it has led to the notion of the ``CSI Effect"  which has altered public understanding of forensic science \cite{maeder2015beyond,Schweitzer2007} and thus has had a societal impact.  CSI has also been the subject of much investigation in media studies because of its importance.

The remainder of the paper is structured as follows: In Section~\ref{sec:related_work}, we discuss some of the seminal related work and in Section~\ref{sec:data}, we discuss the CSI dataset used throughout this study and describe the data manipulation and augmentation techniques we performed. Section~\ref{sec:setup} describes our experiments and  discusses the results of these experiments. Finally, in Section~\ref{sec:conclusion}, we make some closing remarks and discuss future directions.

It is important for the reader to note that
this work does not include any direct comparison to the work of others because there is nothing for us to compare against.
Thus a direct evaluation of this work is not possible because of the nature of the topic. Instead, our novel findings present  the kinds of insights which can be gleaned from computational memorability analysis of visual content and this points to the promise that such analysis can have in areas like educational content, marketing, film and TV production, and content retrieval.

\section{Related Work}
\label{sec:related_work}
\subsection{Media Memorability}
Video Memorability (VM) is a natural progression from a related research discipline: \emph{image memorability} (IM). The work of \cite{Isola2011} is regarded as a seminal paper on IM in which it is stated that memorability is an intrinsic property of an image, an abstract concept similar to image aesthetics which can be computed automatically using modern image analysis techniques \cite{hu2018image}. The use of deep-learning frameworks has led to results at near human levels of consistency in the task of IM prediction \cite{Baveye2016,Fajtl2018,Akagunduz2020}  partly  attributed to the development of image datasets with annotated memorability scores \cite{Kholsa2015}. 
In contrast, VM is still in its early stages and only recently has there been  available data sets \cite{Cohendet2019,Newman2020} on which to train and test VM models.

The Mediaeval Predicting Media Memorability task first took place in 2018, has run annually since then \cite{Mediaeval2018,Mediaeval2019,Mediaeval2020,Mediaeval2021,kiziltepe2021annotated} and  has been  central to the  development of VM. The 2021 edition saw the use of vision transformers \cite{Dosovitskiy2021,Bao2021,Radford2021} across all participant submissions. The  focus of work in the area is predominantly on developing memorability prediction techniques and to date the most applicable use of VM  has been in the analysis of TV advertisements  \cite{Mai2009,Shen2020}.

\subsection{Affective Video Content Analysis}
Affective Video Content Analysis (AVCA) is a research area closely related to VM which investigates the emotions elicited by video among its viewers \cite{Baveye2018}. Like VM, AVCA-based research aims to generate metacognitive insights capable of improving video content indexing and retrieval \cite{Hau2005,Zhao2011}. Besides this, AVCA research could, for example,  protect children from emotionally harmful media \cite{Wang2011}, as well as create mood-dependent video recommendation \cite{Hanjalic2006}. Past research in this domain has shown the modelling of some signatures of film such as tempo \cite{Adams2000}, violence \cite{Penet2012,Eyben2013}, and emotion \cite{Sun2009}.

\subsection{Macabre Fascination in CSI: The Rise of the Corpse}
\label{subsec:corpse-porn}
The impact of crime-drama TV series such as CSI on modern pop-culture stretches far beyond academic research,  even resulting in the coining of a new term: `The CSI Effect' \cite{Schweitzer2007,maeder2015beyond}, mentioned earlier. Forensic autopsy portrayed in CSI acts as a softening lens in which exploration of the dead who are usually the victims of the crime which is the focus of the episode, is socially-palatable entertainment \cite{Penfold2016}. As CSI viewership increased from the first series in 2001, so did the number of forensic science courses available across educational bodies\footnote{Retrieved from: \url{http://www.telegraph.co.uk/education/universityeducation/3243086/CSI-leads-toincrease-in-forensic-science-courses.html}}. Early 2000s crime-drama TV such as CSI sexualises the cadavers present in their series through the ever-present ``beautiful female victim'' motif. The sexualised corpse is a stock character that makes regular appearances. Our eyes are invited to linger on the passive, beautiful, dead bodies that lay in the morgue. This phenomenon captured the public's attention and increased the abundance of sexualised cadavers throughout subsequent seasons of CSI while other series such as \textit{Law \& Order} began adding autopsies to their shows  \cite{Foltyn2008}.

\section{The CSI Dataset}
\label{sec:data}
\subsection{Source Data}
We use 39 annotated episodes of CSI in our analysis, curated by Frermann {\em et al.} \cite{Frermann_Cohen_Lapata_2018} and  available on the Edinburgh NLP GitHub\footnote{The repository hosting the original CSI annotation data is available at \url{https://github.com/EdinburghNLP/csi-corpus} and the memorability scores we computed on this corpus are available at \url{https://github.com/scummins00/CSI-Memorability/blob/main/data/shot_memorability.csv}}. The original source data consists of two corpora, the first of which is {\em Perpetrator Identification (PI)}.  PI corpus' files are annotated at word-level for each episode. A word can be uttered by a speaking character or is part of the screenplay. Each word is accompanied by \emph{case ID}, \emph{sentence ID}, \emph{speaker}, \emph{sentence start time}, \emph{sentence end time}. We also have binary indicators such as \emph{killer gold}, \emph{suspect gold}, and \emph{other gold} which indicate whether the word uttered mentions the killer, suspect, or otherwise.

The second corpus in the original annotated data set is {\em Screenplay Summarisation (SS)}  presented at scene-level (a section of the story with its own unique combination of setting, character, and dialogue). For a given episode, the  dataset consists of each scene annotated with \emph{Scene ID}, \emph{screenplay}, as well as the scene \emph{aspect}. Aspect is a categorical feature  indicating the scene's significance and  can be one or many of the following:
{\textit{Crime Scene}}, {\textit{Victim}}, {\textit{Death Cause}}, {\textit{Evidence}}, {\textit{Perpetrator}}, {\textit{Motive}}, or {\textit{None}}.

Each of the 39  episodes  was  also available to us in video format by scraping directly from the DVDs purchased from Amazon, as well as an extra 3 episode video files. 

\subsection{Data Preprocessing}
\subsubsection{Data Manipulation:}

We aggregated each of our PI episode datasets from word-level to sentence-level, combining the \emph{killer gold}, \emph{suspect gold}, and \emph{other gold} binary indicators previously mentioned into a single categorical column. We then disaggregated our SS corpora from scene-level to sentence-level. With  equivalent datasets of the same configuration, we then merged them and Table~\ref{tab:augmented_data} shows a section of one of our augmented and merged datasets for S1.E8.

\begin{table*}[t]
\centering
\caption{Sample augmented dataset from Season 1, Ep. 8. Original data from \cite{Frermann_Cohen_Lapata_2018}.}
\label{tab:augmented_data}
        \begin{tabular}{cccccclc}
        \toprule 
        caseID & sentID & speaker & type mentioned & start & end & sentence & aspect\\
        \midrule 
        1 & 6 & Grissom & other & 00:36.5 & 00:41.5 & where's the girl? & Victim\\
        1 & 7 & Officer & other & 00:41.5 & 00:46.6 & she's down this hall & Victim\\
        1 & 8 & None & None & 00:46.6 & 00:51.7 & Grissom turns the corner & Crime scene\\
        1 & 9 & None & None & 00:51.7 & 00:56.8 & the officer signals Grissom inside & Crime scene\\
        \bottomrule
        \end{tabular}
\end{table*}

\subsubsection{Data Augmentation:}
Features available from the annotation are rich and allow us to analyse data in a variety of ways. However, we noted two shortcomings: (i) the data provides little context regarding suspects throughout an episode other than the \textit{type mentioned} feature and (ii) the \textit{aspect} feature does not allow us to identify scenes involving suspects. Also, a portion of scenes involving the killer are not labelled with the `Perpetrator' aspect.
We extended the \textit{aspect} feature of any sentence in a scene in which the perpetrator or a suspect speaks so as to have the values of `Perpetrator' or `Suspect', respectively.

\subsection{Pre-processing of Video Files }
\label{subsec:video_manipulation}
We subdivided our clips into shots which allows for more specific and fine-grained indexing, in turn reducing the amount of video we needed to process per analysis. We used a neural shot-boundary-detection (SBD) framework, \emph{TransNet V2},\footnote{TransNet V2 is an SBD framework: \url{https://github.com/soCzech/TransNetV2}.} described in \cite{transnetV2}, to segment  episodes into individual shots.

\subsection{Data Generation}

\begin{figure*}[ht]
    \centering
    \includegraphics[width=\textwidth]{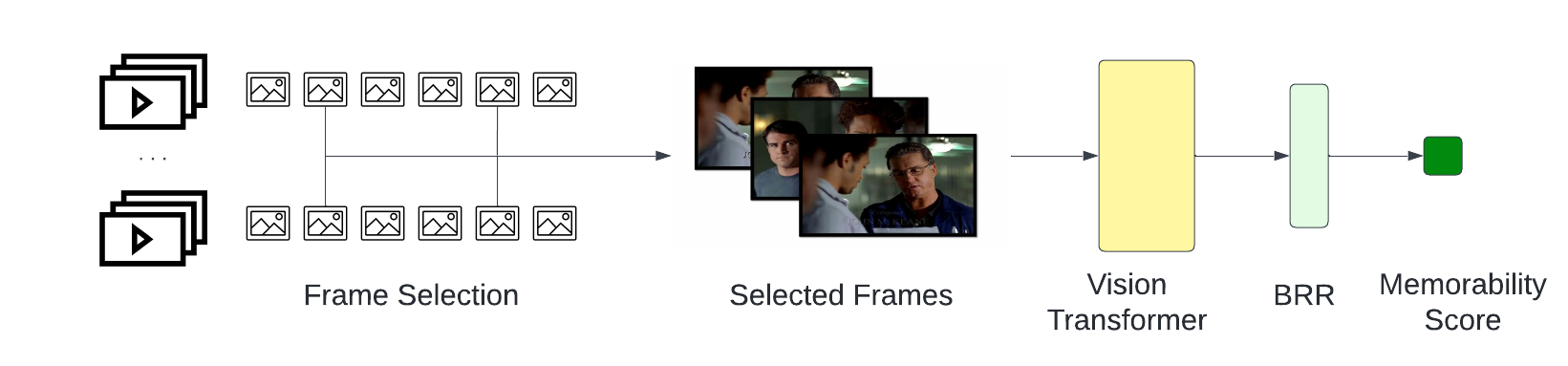}
    \caption{Our memorability prediction framework, extracting representative frames  and computing a memorability score for each  shot. \label{fig:framework}}
\end{figure*}

The use of vision transformers in predicting memorability scores is shown in \cite{Constantin2021,Kleinlein2021}. Our vision transformer architecture is the CLIP \cite{Radford2021} pre-trained image encoder. We use this encoder to train a Bayesian Ridge Regressor (BRR) on the Memento10k dataset \cite{Newman2020}. BRR was used previously by \cite{Sweeney2021} for computing memorability scores based on extracted features. For each training sample, the encoder extracts features from frames sampled at a rate of 3 frames per second. The BRR is then trained on the features and memorability score pairs. To compute the memorability of unseen CSI clips, we first extract representative frames using a temporal-based frame extraction mechanism. These are used as input to the vision transformer architecture to extract features. Our BRR model then produces a memorability score based on these features. This framework is shown in Figure~\ref{fig:framework}.

\section{Methodology and Experimental Results}
\label{sec:setup}
\label{sec:results}
We now present the distribution of memorability scores across each episode, relating it to its  metadata from  annotations taken from \cite{Frermann_Cohen_Lapata_2018}.

For each episode, we plotted the memorability scores associated with shots in chronological order  to create a time-series signal representing the intrinsic memorability of the episode. We then annotated each episode's signal according to various metadata including (i) the scene aspect, and (ii) the character speaking (including None). We used Laplacian smoothing described in Equation~\ref{eq:smooth}  to remove the jaggedness present in raw memorability scores 
\begin{equation} \label{eq:smooth}
    \hat{x}_{i}=\frac{1}{N}\sum_{j=1}^{N}\hat{x}_{j}
\end{equation}
where $N$ is the size of the smoothing window. We generated signals for each  episode using a range of smoothing window sizes from  15 to  305 shots.

\begin{figure*}[ht]
    \centering
    \subfigure[]{\includegraphics[width=0.39\linewidth]{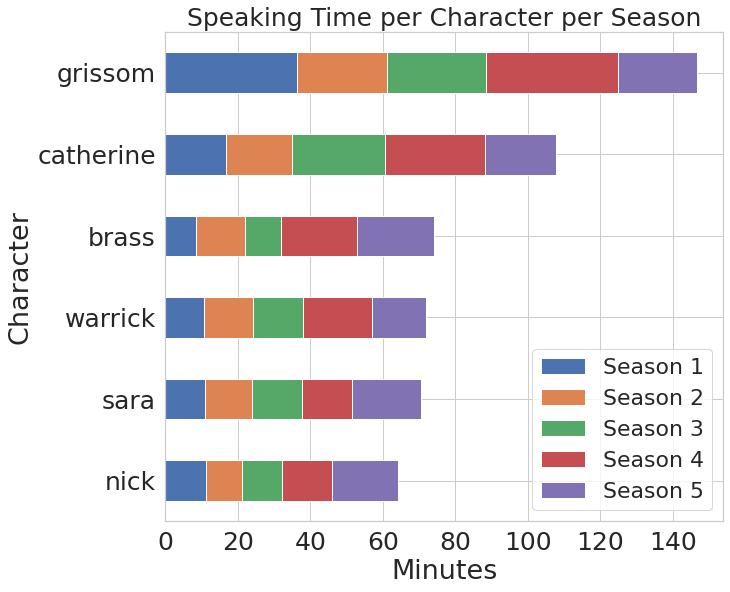}}
    \hspace{2cm}
    \subfigure[]{\includegraphics[width=0.39\linewidth]{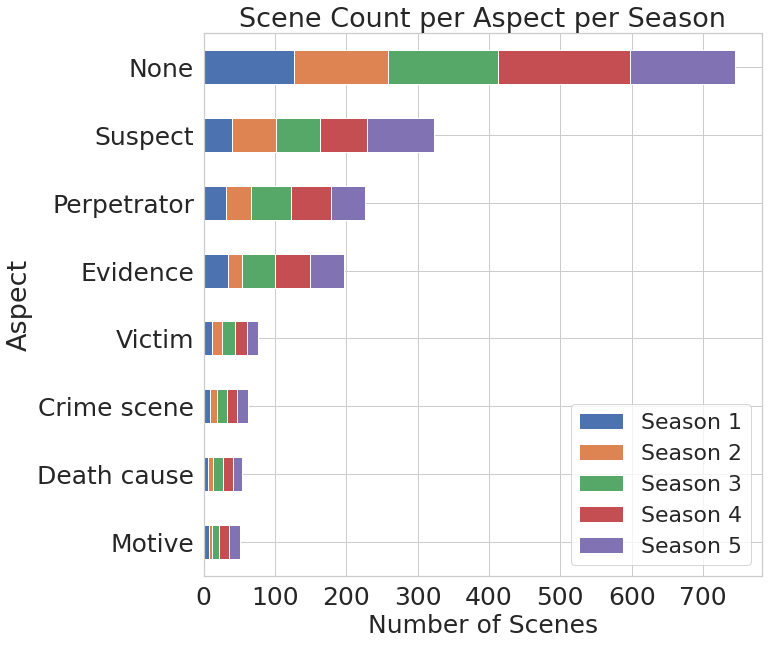}}
    \caption{(a) The time (mins) spent speaking by each main cast member. (b) The number of scenes in which each aspect value is present. \label{fig:statistics_prelim}}
\end{figure*}

To compliment the shot memorability score distributions we gathered memorability scores across the distribution of characters, and of aspects. By aggregating the overall speaking time of the main characters, we identify the lead  as well as the secondary characters. We performed a similar analysis to understand the scene count per aspect value. Our results are shown in Figure~\ref{fig:statistics_prelim}.

\begin{figure*}[htb]
    \centering
    \subfigure[]{\includegraphics[width=0.65\linewidth]{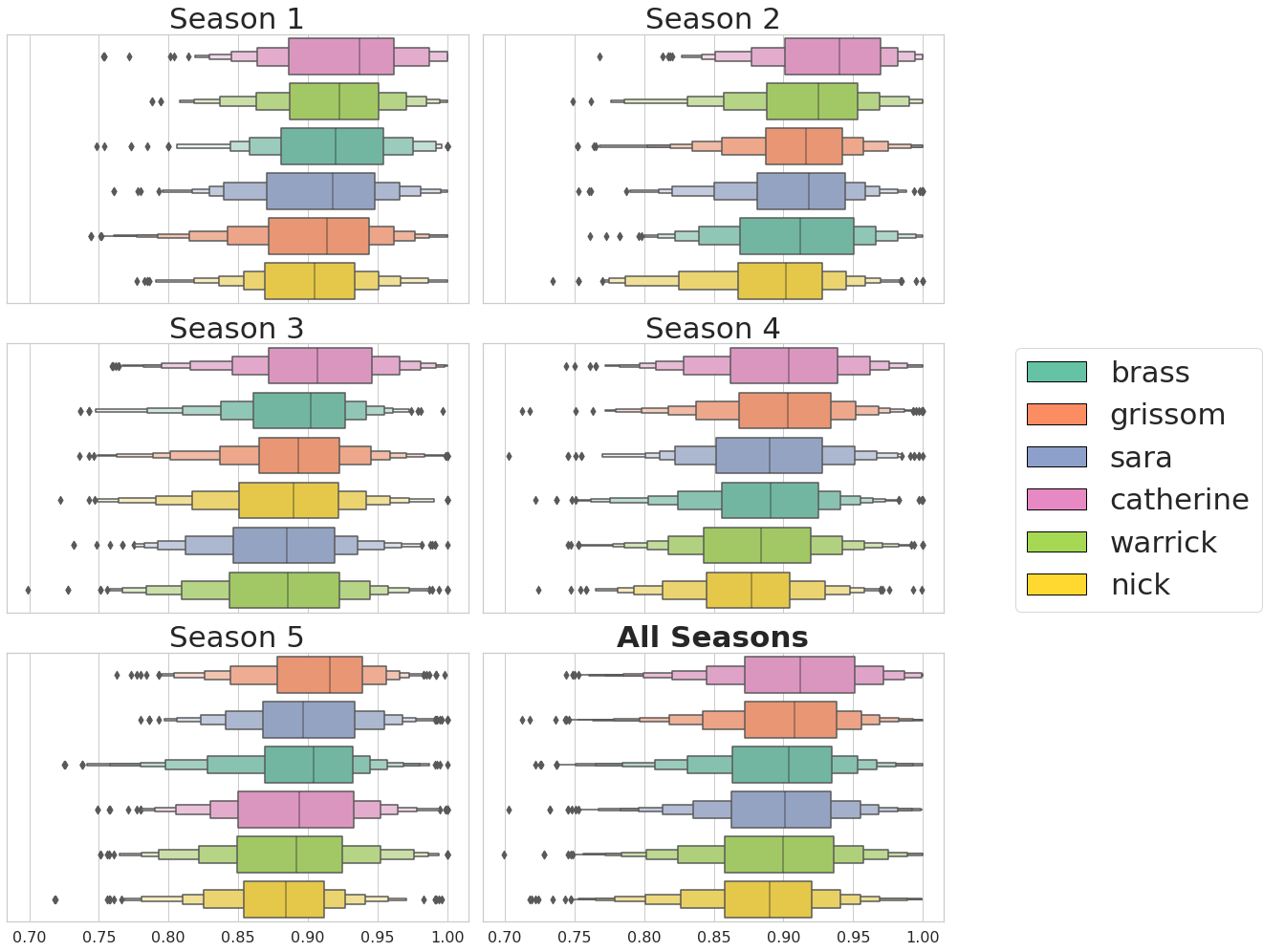}}
    \subfigure[]{\includegraphics[width=0.65\linewidth]{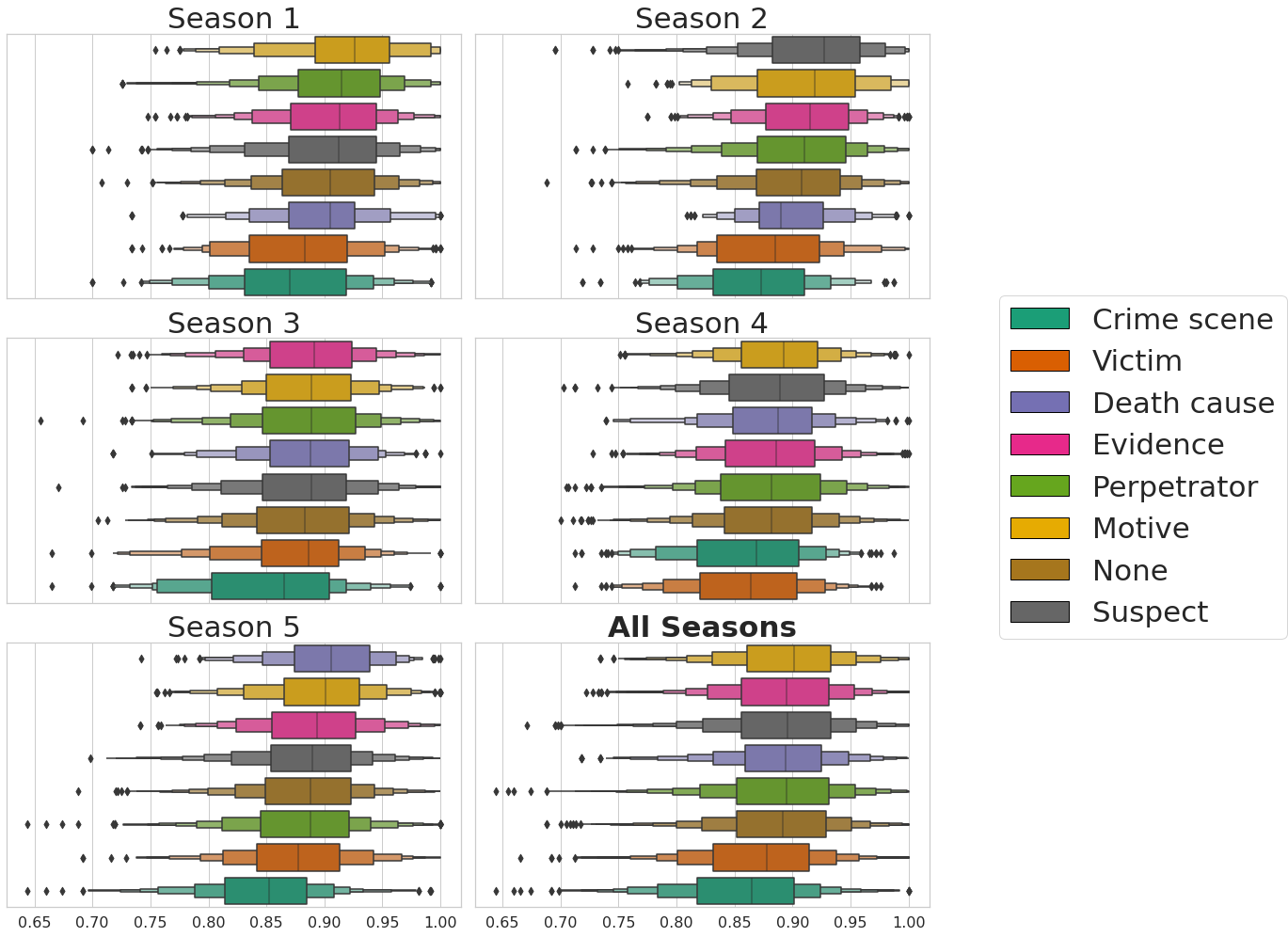}}
    \caption{Graphs display results from Seasons 1 to  5 and then results for  all seasons combined. 
    (a) displays computed memorability scores for shots with the main characters, while (b) displays computed memorability scores for shots with associated with aspect values.}
    \label{fig:boxens}
\end{figure*}

We investigated the memorability distribution associated with the series characters and aspects and this is presented in Figure~\ref{fig:boxens} parts (a) and (b) in each individual season as well as across all 5 seasons. We investigated: i)the main character speaking, and (ii) the scene aspect.
The lack of variation across Figure~\ref{fig:boxens} (a) and (b) and the relatively high memorability scores is an artefact of the Memento10k memorability dataset \cite{Newman2020} used to train the memorability model we used where we rarely see a video segment given a memorability score below 0.7. Instead we are  interested in repeating patterns   e.g. which parts of  shows consistently result in higher or lower memorability scores and what  these patterns might reveal to us about the series.

In Figure~\ref{fig:boxens} (a), we see that ordering the cast in terms of memorability over all 5 seasons is almost identical to that seen in Figure~\ref{fig:statistics_prelim} (a). Catherine has less speaking time than Grissom, but is consistently considered more memorable. With the exception of Season 3, Nick is simultaneously the least-memorable character and the character with the least time spent speaking. We observe a somewhat positive linear relationship between a character's importance (in terms of screen-time), and their memorability.

In Figure~\ref{fig:statistics_prelim} (b), the proportion of scenes in which each aspect value appears correlates with the purpose in which that aspect serves. For example, `Motive' appears  least across all seasons as only a small portion near the end of each episode is dedicated to revealing a perpetrator's motive. Similarly, `Crime scene' and `Victim' scenes usually occur near the beginning of an episode. Despite this, as  shown in Figure~\ref{fig:boxens} (b), `Motive' scenes are considered the most memorable  across all 5 seasons. In  contrast, both `Crime scene' and `Victim' scenes are considered the least memorable.

Scenes described with the `Death Cause' aspect value are associated with autopsy scenes in which viewers are shown flashbacks to the crime committed, as well as the cadaver of a usually youthful, attractive victim. We discussed the importance of these scenes and their pivotal role in the development of TV crime series in Section~\ref{subsec:corpse-porn}. In Figure~\ref{fig:boxens} (b), the intrinsic memorability of this scene type increases as the seasons progress. These pedagogic, grotesque, scenes capture the audience's fascination because of their brash autoptic vision \cite{Tait2006} and are a staple of the crime-drama TV genre and contribute to its popular success. The visual memorability of these scenes correlates with the success of this scene-type.

\section{Conclusions and Future Work}
\label{sec:conclusion}

Video memorability is a recent computational tool for analysis of visual media  which is more abstract and higher-level than raw content analysis like object detection or action classification. In this paper we have used the computation of video memorability at video shot level to analyse a popular procedural crime-drama TV series revealing insights not previously visible.

Our investigation creates a mélange of intertwined arguments ranging from more statistically-driven interpretations of film such as cinemetrics, to artistic studies on the relationship between a series' popularity and its most memorable moments. We hypothesise that scene and character importance is related to the memorability of a scene, and show that correlation exists between factors of the show not previously visible. We show that these correlations are not random, but instead have complex interpretation. The memorability scores we generated uncovered patterns within the early seasons of CSI, highlighting the significance and memorability of an episode's finale in comparison to other scenes. Our memorability scores for lead cast members heavily correlate with the characters' importance, considering on-screen time as importance. The work here is one of the first efforts in the application of memorability scores beyond benchmarking environments.

Our first area for future work is to increase the data size by including more episodes of CSI, as well as episodes from other shows such as \textit{CSI: Miami} or other `Whodunnit-styled' series such as \textit{Law \& Order}.
In our study, we used a temporal-based frame extraction method to extract representative frames from shots. This could be enhanced via Representative Frame Extraction techniques \cite{Shruthi2017}, which are capable of automatically selecting the most representative frames from video content.

It is clear that we have developed an importance-proxy by analysing various metadata from CSI. Ideally, we would define a user task in which participants' genuine memorability of aspects of a crime-drama series can be examined. This gives a real measure of importance with respect to the show, rather than an importance proxy.

The  novel findings we have presented here reveal insights  from computational memorability scores for video shots  of a TV crime series.  As mentioned earlier, this is useful and interesting from a film or TV studies perspective but it also  shows the potential that computational memorability has in other areas.
We could now analyse or even edit videos to be used in online education so as to maximise the memorability of key moments.  We could  structure video content used in marketing and advertising, or as we have seen here in film and TV production, so that key messages are delivered with maximised memorability. Future work in these and other areas may indeed exploit computational memorability as a criterion for composing video content.

\begin{acks}
This publication has emanated from research partly supported by Science Foundation Ireland (SFI) under Grant Number SFI/12/RC/ 2289\_P2 (Insight SFI Research Centre for Data Analytics).
\end{acks}

\bibliographystyle{ACM-Reference-Format}
\bibliography{CBMI-2022}

\end{document}